\documentclass[journal]{IEEEtran}
 \usepackage{graphicx}
 \DeclareGraphicsExtensions{.pdf,.jpeg,.png}

\usepackage[cmex10]{amsmath}
\usepackage{multirow}

\usepackage{bm}
\usepackage[normalem]{ulem}

\usepackage{array}
\usepackage{mdwmath}
\usepackage{mdwtab}
\usepackage{eqparbox}
\usepackage{amsfonts}
\usepackage{color}

\usepackage{cite}
\usepackage{enumerate}
\usepackage[tight,footnotesize]{subfigure}

\usepackage{xcolor}

\ifCLASSINFOpdf
\else
\fi
\begin{document}
%
\title{Attention to Head Locations for Crowd Counting}
%
%
%

\author{Youmei Zhang,~
        Chunluan Zhou,~
        Faliang Chang,~
        and Alex C. Kot,~\IEEEmembership{Fellow Member,~IEEE}
\thanks{Manuscript received March 1st, 2018; revised xx xx, xx. This work was supported in part by the National Natural Science
Foundation of China under Grant61673244, Grant61273277 and Grant61703240) and was carried out at the Rapid-Rich Object Search (ROSE) Lab at the Nanyang Technological University, Singapore. The ROSE Lab is supported by the Infocomm Media Development Authority, Singapore. \emph{(Corresponding Author: Faliang Chang)}}
\thanks{Y. Zhang and F. Chang are with the School of Control Science and Engineering, Shandong University, Jinan, China 250061
(e-mail: zym5289@gmail.com; flchang@sdu.edu.cn).}
\thanks{C. Zhou and A. C. Kot are with the School of Electrical and Electronic Engineering, Nanyang Technological University, Singapore
639798 (e-mail:czhou002@e.ntu.edu.sg; eackot@ntu.edu.sg).}}

%
%

%



\maketitle

\begin{abstract}
Occlusions, complex backgrounds, scale variations and non-uniform distributions present great challenges for crowd counting in practical applications. In this paper, we propose a novel method using an attention model to exploit head locations which are the most important cue for crowd counting.
The attention model estimates a probability map in which high probabilities indicate locations where heads are likely to be present.
The estimated probability map is used to suppress non-head regions in feature maps from several multi-scale feature extraction branches of a convolutional neural network for crowd density estimation,
which makes our method robust to complex backgrounds, scale variations and non-uniform distributions.
In addition, we introduce a relative deviation loss to compensate a commonly used training loss, Euclidean distance, to improve the accuracy of sparse crowd density estimation.
Experiments on ShanghaiTech, UCF\uline{~}CC\uline{~}50 and WorldExpo'10 datasets demonstrate the effectiveness of our method.
\end{abstract}

\begin{IEEEkeywords}
Crowd Counting, Convolutional Neural Network, Head Locations, Attention Model, Relative Deviation Loss.
\end{IEEEkeywords}

%
\IEEEpeerreviewmaketitle


\section{Introduction}

\begin{figure}[!t]
\centering{
\subfigure[Occlusion]{\includegraphics[width=1.5in]{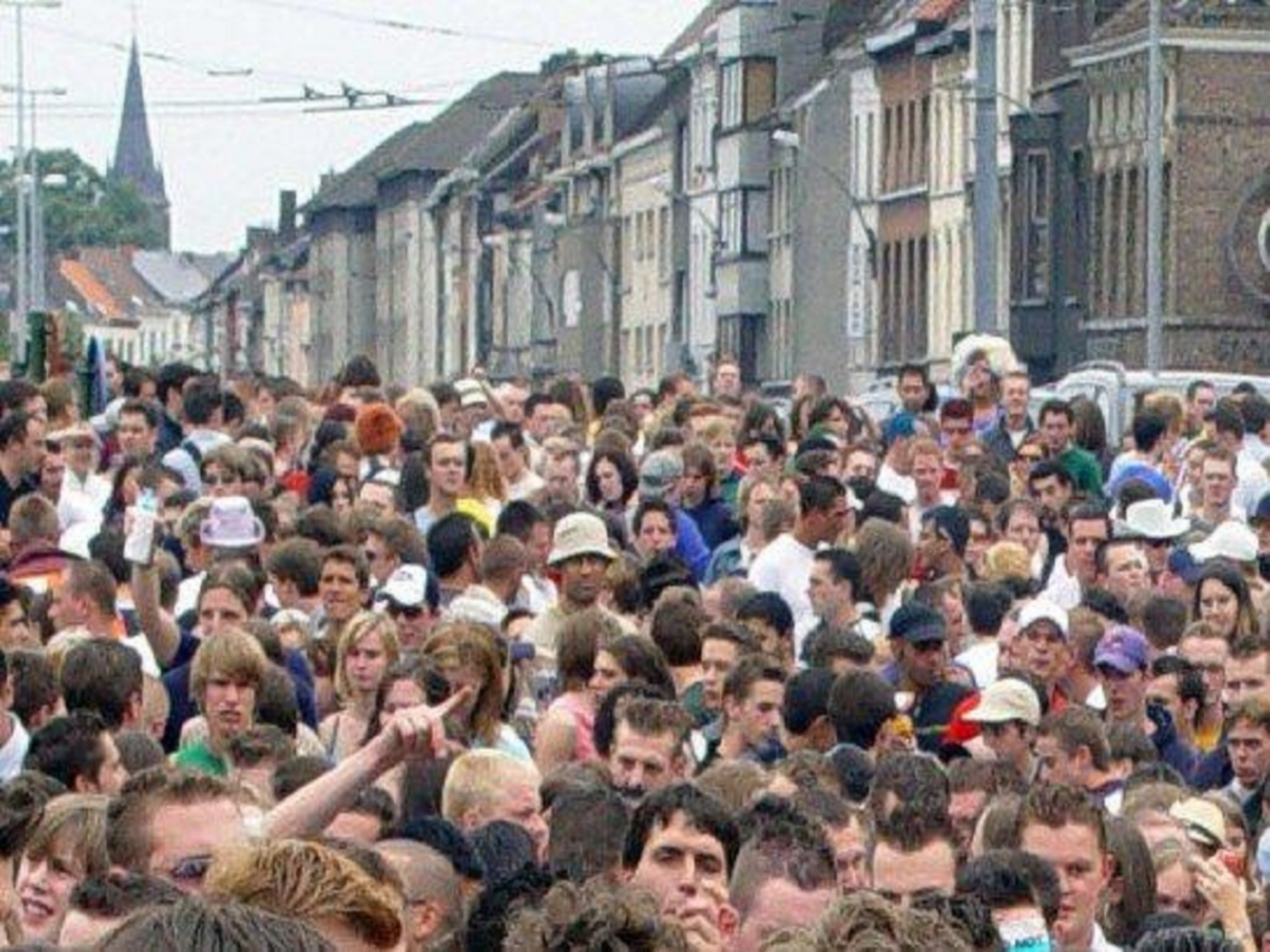}%
\label{COcclusion}} \hfil
\subfigure[Complex background]{\includegraphics[width=1.5in]{Cbackground.pdf}%
\label{background}}\hfil\\
\subfigure[Scale variation]{\includegraphics[width=1.5in]{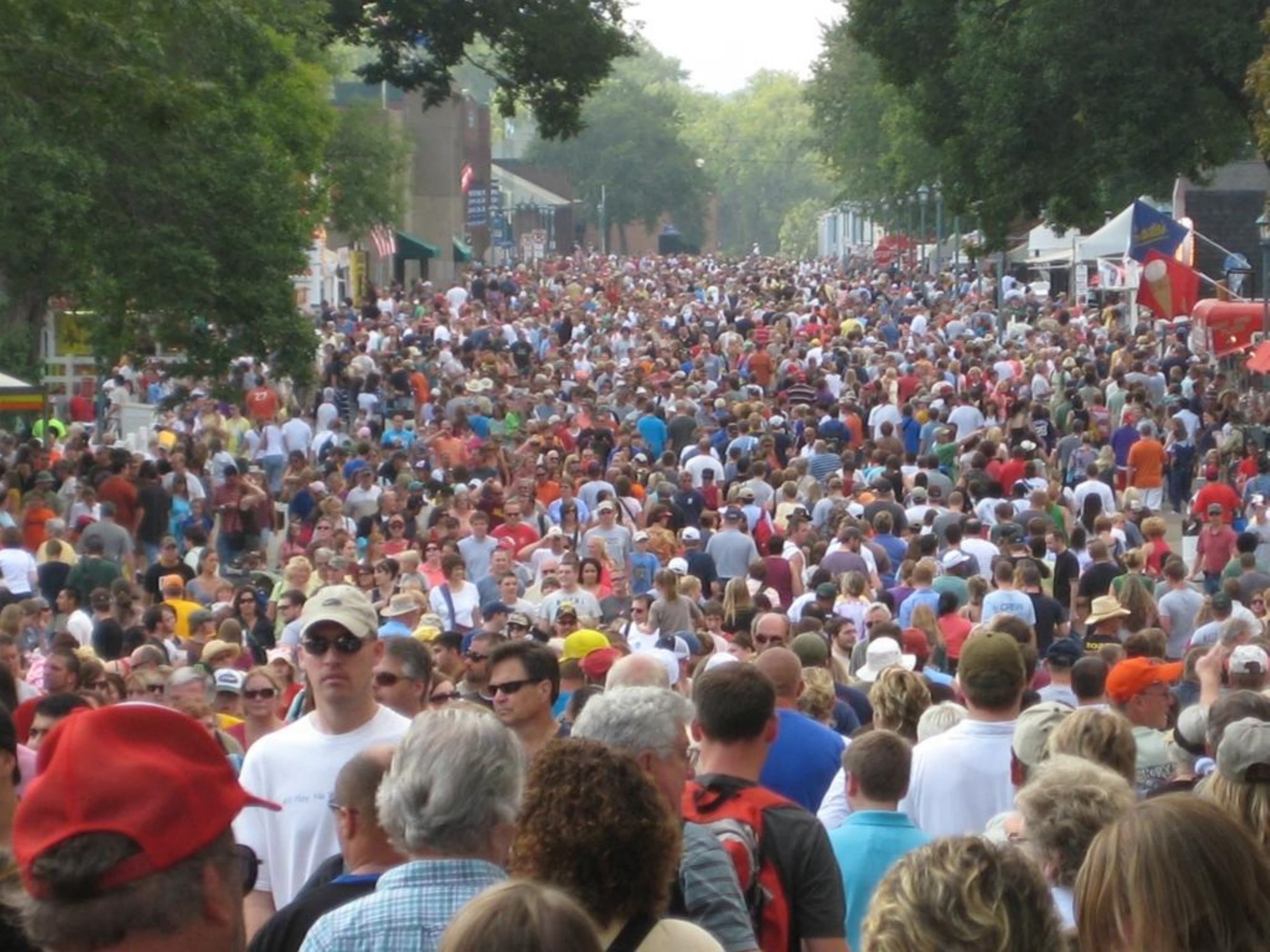}%
\label{Scale}} \hfil
\subfigure[Non-uniform distribution]{\includegraphics[width=1.5in]{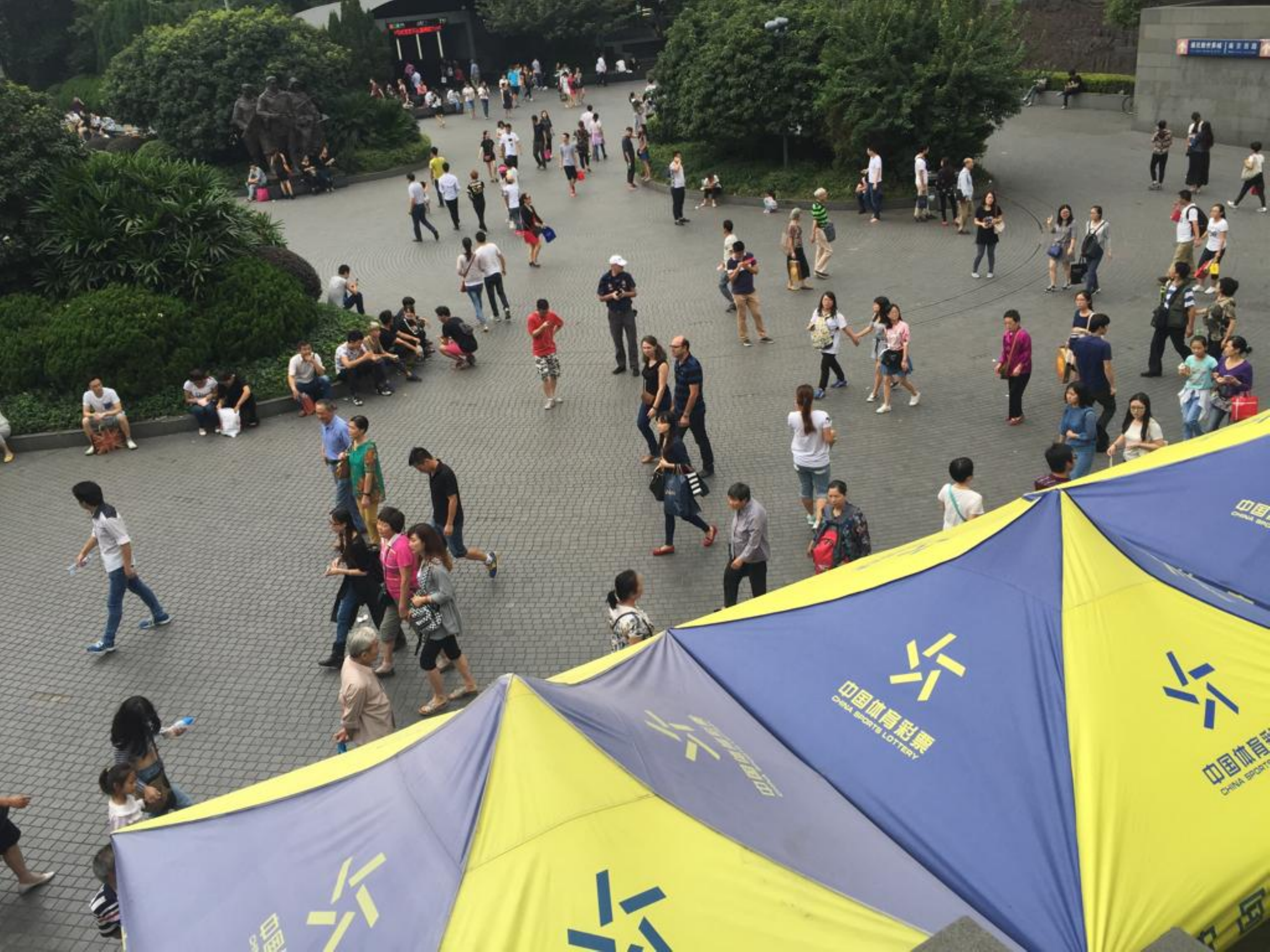}%
\label{distribution}}}
\caption{Challenges for crowd counting.}
\label{fig_challenge}
\end{figure}

\IEEEPARstart{W}{ith} increasing demands for intelligent video surveillance, public safety and urban planning,
improving scene analysis technologies becomes pressing \cite{li2015crowded,zhang2016data}.
As an important task of scene analysis, crowd counting has gained more and more attention from multimedia and computer vision communities in recent years for its applications such as crowd control, traffic monitoring and public safety.
However, the crowd counting task comes with many challenges such as occlusions, complex backgrounds,
non-uniform distributions and variations in scale and perspective \cite{sindagi2017survey}, as Fig. \ref{fig_challenge} shows.
Many algorithms have been proposed to address these challenges and increase the accuracy of
crowd counting \cite{zhang2015cross,zhang2016single,sindagi2017cnn,sindagi2017generating}.

Recent methods based on convolutional neural networks (CNNs) have achieved a significant improvement in crowd counting \cite{sindagi2017survey}.
A multi-column CNN (MCNN) is proposed in \cite{zhang2016single} to address the scale-variation problem
by using several CNN branches with different receptive fields to extract multi-scale features.
A cascaded CNN \cite{sindagi2017cnn} learns high-level prior which is incorporated into the crowd density estimation branch
of the CNN to boost the performance.
In \cite{sindagi2017generating}, both global and local context are exploited to generate high-quality crowd density maps.
Despite these methods have achieved promising performance,
they neglect two aspects which could be exploited to further improve the accuracy of crowd counting.
Firstly, these methods do not well exploited head locations in images which are the most important cue for crowd counting.
Actually, head locations are usually used to generate ground-truth density maps in crowd counting datasets,
e.g. ShanghaiTech \cite{zhang2015cross} and UCF\uline{~}CC\uline{~}50 \cite{idrees2013multi} datasets.
Although the generated ground-truth density maps from head locations are used to learn a CNN for regression,
these methods do not explicitly give more attention to head regions during training and testing.
In other words, they treat head and background regions equally.
Secondly, the network training in these methods are dominated by dense crowd examples because of
the use of the Euclidean distance between ground-truth and estimated density maps as the training loss.
Generally, it is much more difficult to predict density maps for dense crowd examples than for spare crowd examples,
leading to far larger training loss for the former.
As a result, sparse crowd examples tend to receive insufficient treatment during training.
However, sparse crowd counting could also be very important for some specific applications.
For instance, in markets and street advertising scenarios, people may be attracted by some commodities and stroll in front of them,
thus forming some sparse crowds.
Counting the number of people in these scenarios to obtain the distributions of crowds could provide useful information regarding
the preferences of customers for businesses and advertisers.


In this paper, we propose a novel method to address the above-mentioned two limitations of existing CNN based approaches.
Fig. \ref{fig_AMCNN} shows the network architecture used in the proposed method.
We incorporate an attention model into the MCNN \cite{zhang2016single} to guide the network to focus on head locations during training and testing.
Specifically, the attention model learns a probability map in which high probabilities indicate locations where heads are likely to be present.
This probability map is used to suppress non-head regions in feature maps from multi-scale feature extraction branches of the MCNN.
In addition, to obtain better density maps for sparse crowds, we also introduces a relative deviation loss which is combined with the commonly used Euclidean loss to train the network of our method.
The relative deviation loss increases the importance of sparse crowd examples during training such that the network is learned to better predict density maps for sparse crowd examples.
We validate the effectiveness of the proposed method on three datasets, ShanghaiTech \cite{zhang2016single}, UCF\uline{~}CC\uline{~}50 \cite{idrees2013multi} and WorldExpo'10 \cite{zhang2015cross}.


The main contributions of this work are summarized as follows:
\begin{enumerate}[1)]
\item To our best knowledge, we make the first attempt to use an attention model for crowd counting.
By incorporating the attention model into the CNN, the proposed method can filter most of background regions and body parts,
therefore improving its robustness to complex backgrounds and non-uniform distributions.
\item  The proposed method is robust to variations in scale because of the use of multi-scale feature extraction branches and the capability of the attention model to locate heads of different sizes.
\item The relative deviation loss is introduced to compensate the Euclidean loss,
therefore improving the accuracy of predicting density maps for sparse crowd examples.
\end{enumerate}




The remainder of the paper is organized as follows.
Section \ref{sec:related work} presents some related works about crowd counting and the attention model.
In Section \ref{sec:proposed method}, our proposed attention model convolutional neural network (AM-CNN) is introduced.
The implementation details are presented in Section \ref{sec:implementation}.
Experimental results are given and discussed in Section \ref{sec:experiment}.
Finally, Section \ref{conclusion} concludes the paper.

\section{Related Work}
\label{sec:related work}
\textbf{Traditional counting methods}: Traditional counting methods can be categorized into detection-based approaches,
regression-based approaches and density estimation-based approaches \cite{sindagi2017survey}.
Detection-based approaches typically estimate the number of people based on detecting objects in the scene with a sliding window \cite{dollar2012pedestrian}.
The object detector is usually a classifier which trained on features such as histogram oriented
gradients \cite{dalal2005histograms} and Haar wavelets \cite{viola2004robust}.
Despite the great success in sparse crowd counting, these methods do not work well when it comes to dense crowds.
Although \cite{felzenszwalb2010object} presents a part-based detection method to cater to this problem, the counting results still remain unsatisfactory.

To overcome the defect of detection-based approaches for dense crowd counting,
some researchers \cite{chen2012feature,chen2013cumulative,ryan2009crowd} attempt to estimate the number of people by regression.
Regression-based approaches aim to find a mapping function between extracted features and the global or local counts.
Typically, global or local features extracted from the image are firstly used to encode some low-level information.
Then the mapping between these features and the counts are learned by a regression model.
To utilize more information to get higher accuracy for dense crowd examples,
Idress et al. \cite{idrees2013multi} combined multiple sources such as low confidence head detections
and repetition of texture elements to count at patch level and then used enforced smoothness constraint to produce better estimates at image level.
Besides, they created a new dataset on a scale that was never tackled before.

For some specific occasions, such as markets, it is more important to estimate the crowd distribution rather than only getting the number of customers.
Therefore, getting the density maps while predicting the counts is of great significance.
Counting approach in \cite{lempitsky2010learning} predicts the density maps based on linear function
 and introduces a new loss (Maximum Excess SubArray, MESA) to increase the counting accuracy.
Pham et al. \cite{pham2015count} propose to use non-linear function to learn the mapping.
Besides, they exploit a crowdedness prior and train two different forests to address large variations in appearance.

\textbf{CNN-based counting methods:} 
CNN-based counting approaches have become the main tend for its great success in various computer vision tasks.
Early CNN-based methods \cite{hu2016dense,zhang2018auxiliary,kumagai2017mixture} predict the number of objects instead of density map.
Hu et al. \cite{hu2016dense} exploit a density level classification task to enrich the features, therefore increasing the counting accuracy.
Similarly, method in \cite{zhang2018auxiliary} classifies the appearance of the crowds while estimating the counts,
which forms auxiliary CNN for crowd counting.
Authors of \cite{kumagai2017mixture} address the appearance change problem by multiplying appearance-weights
 output by a gating CNN to a mixture of expert CNNs.
As aforementioned, estimating the crowd distribution while getting the counts is more applicable in some specific scenarios.
Therefore, some researchers attempt to get the counts by density map prediction based on CNN architectures.

Zhang et al. make their first attempt to address the challenge of complex backgrounds by utilizing CNN to estimate the density map,
 which also denotes the counts by the sum of pixel values.
To make use of both high-level semantic information and low-level features, Boominathan et al.
\cite{boominathan2016crowdnet} makes a combination of deep and shallow, fully convolutional network to estimate the density maps.
Some algorithms \cite{zhang2016single,shang2016end,onoro2016towards,sam2017switching} are proposed to cater to large variations in scale and perspective.
The MCNN \cite{zhang2016single} presents several CNN branches with different receptive fields, which could extract multi-scale features and enhance
the robustness to large variations in people/head size.
The Hydra CNN in \cite{onoro2016towards} provides a scale-aware solution to generate the density maps by training the regressor
with a pyramid of image patches at multiple scales.
To make full use of sharing computations and contextual information,
local and global information are leveraged in \cite{shang2016end} by learning the counts of both local regions and overall image.
Authors of \cite{sam2017switching} 
propose a switching-CNN by adding a switch to the MCNN \cite{zhang2016single}.
They utilize an improved version of  VGG-16 as the switch classifier to choose a best CNN regressor for the original image.
Sindagi et al. \cite{sindagi2017generating} aims at generating high quality density maps by
using a Fusion CNN to concatenate features extracted by Global Context Estimator (GCE), Local Context Estimator (LCE) and Density Map Estimator (DME).
In addition, their counting architecture is trained in a Generative Adversarial Network to get shaper density maps.

\textbf{Attention Model:} The attention model has been widely used for varies computer vision tasks,
such as image classification \cite{xiao2015application} and segmentation \cite{chen2016attention},
object detection \cite{zhou2014object} and classification \cite{zhao2017diversified}, action recognition \cite{hou2017content,liu2017skeleton}, pose estimation \cite{chu2017multi}, scene labeling \cite{abdulnabi2017episodic} and video captioning \cite{gao2017video}.
Xiao et al. \cite{xiao2015application} propose a two-level attention for image classification: object-level attention and part-level attention.
The former selects the patches relevant to the task domain while the latter focuses on local discriminate patterns.
The two level attentions compensate each other nicely with late fusion.
Chen et al. \cite{chen2016attention} exploit the attention model to measure the importance of different-scale features after generating multi-resolution inputs for semantic segmentation.
In \cite{hou2017content}, a Content Attention Network (CANet) for action recognition is proposed to improve the robustness to the irrelevant content by addressing the attention mechanism and using clean videos as the guidance for training.
Liu et al.\cite{liu2017skeleton} proposed a Global Context-Aware long short-term memory (GCA-LSTM) network,
which introduces a recurrent attention model to LSTM and thus selectively focusing on the informative joints with regarding to global context information.
Apart from context-aware attention, Chu et al. \cite{chu2017multi} incorporate multi-resolution attention and hierarchical attention into an hourglass network for pose estimation.
Their model can focus on different granularity from local salient regions to global semantic-consistent spaces.
A contextual attention model is utilized in \cite{abdulnabi2017episodic} to assign different power weights to surrounding patches,
 and thus adaptively selecting relevant patches for scene labeling.
In \cite{gao2017video}, Gao et al. integrate LSTM with an attention mechanism that uses the dynamic weighted sum of local two-dimensional convolutional neural network representations to capture salient structures of video, thus  generating sentences with rich semantic content for video captioning.
All of these works have demonstrated that the attention model allows the network to focus on most relevant features as needed.

\section{The Proposed Method}
\label{sec:proposed method}
\begin{figure}[!ht]
\centering
\includegraphics[width=3.0in]{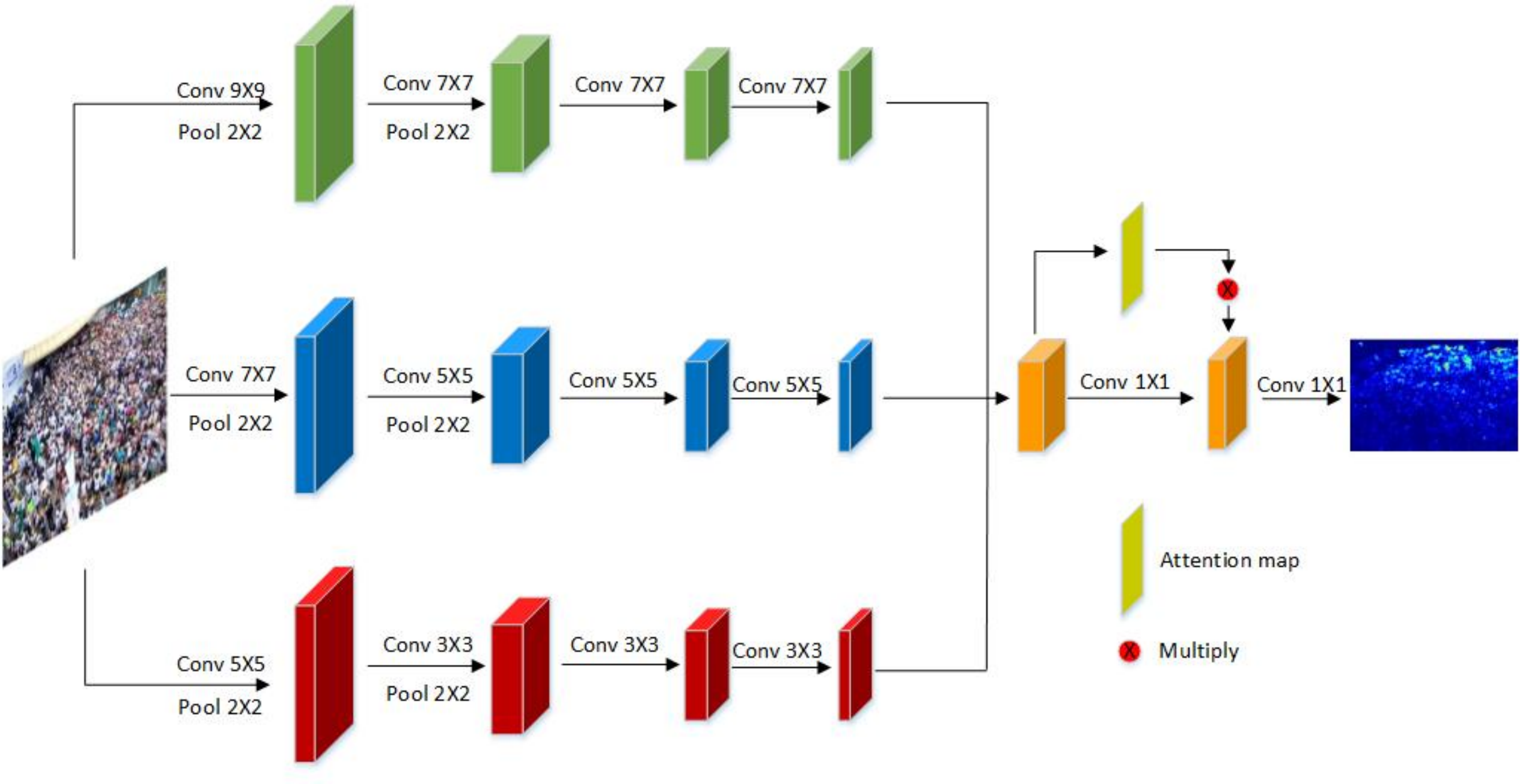}
\caption{Architecture of the AM-CNN.
The image is firstly fed into three shallow CNN branches to extract multi-scale features.
These branches are with different sizes of filters,
which can be represented as large (9-7-7-7), medium (7-5-5-5) and small (5-3-3-3).
Then the feature maps from different branches are concatenated to generate attention features by the attention model.
Since containing 2 max pooling layers in each CNN branches, this architecture finally outputs a density map with 1/4 size of the original image.}
\label{fig_AMCNN}
\end{figure}

The proposed AM-CNN consists of 3 shallow CNN branches and an attention model.
The CNN branches with different receptive fields are firstly exploited to extract multi-scale features.
Then the attention model is incorporated to emphasize head locations
regardless of the complexity of scenes, the non-uniformity of distributions and the variability of scale and perspective.
In addition, a relative deviation loss is used to compensate Euclidean loss during the training process.
The architecture of the proposed AM-CNN is illustrated in Fig. \ref{fig_AMCNN} and discussed in detail as follows.

\subsection{Feature Extraction with multi-receptive fields}
Some of previous works \cite{zhang2016single,sindagi2017generating,onoro2016towards,sam2017switching}
 exploited multi-column networks with different receptive fields to address the variations in scale
 since different sizes of receptive fields can cope with the diversity in object-size \cite{zhou2014object}.
Inspired by successful use of the MCNN \cite{zhang2016single,sindagi2017generating,sam2017switching},
 we select part of it to extract multi-scale features.
The multi-column architecture with larger filter sizes or more columns may cater to larger variations in scale,
 but it brings a time-consuming parameter adjustment task.
Since the proposed method mainly focuses on the effect of the attention model for crowd counting,
 we use the same filter sizes and channels as \cite{zhang2016single} and \cite{sam2017switching}.
But different from them, the multi-column network in this paper is used to generate high-dimensional feature maps
 rather than transforming the input into a density map directly.

\begin{figure*}[!ht]
\centering
\includegraphics[width=5.5in]{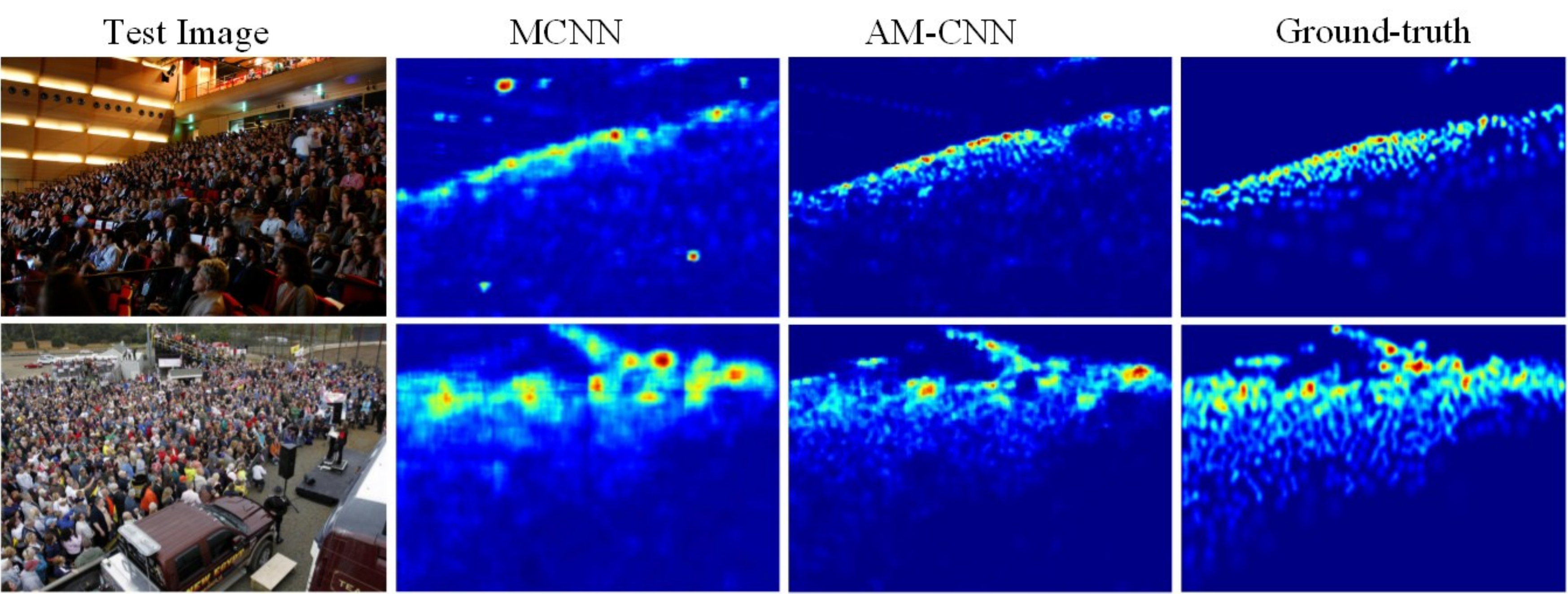}
\caption{Density maps generated by different methods.
Best viewed in color.}
\label{fig_MCNN}
\end{figure*}

Density maps generated by the MCNN \cite{zhang2016single} contain complex backgrounds, which impact the counting accuracy seriously.
In addition, the distinction between large and small objects is not so obvious in density map, as Fig. \ref{fig_MCNN} shows.
The most important cue for crowd counting, head locations, is the key to address the above problems.
Therefore, we need an operation to guide the network to give more attention to head locations and suppress non-head regions.
In virtue of the strong object-focused capability, an attention model is incorporated into the MCNN and
 thus forming a new architecture which could generate more accurate density maps.
We will describe the attention model in Section \ref{sec:AM}.

\subsection{The attention model for crowd counting}
\label{sec:AM}
Visual attention is an essential mechanism of the human brain for understanding scenes effectively \cite{chu2017multi}.
Therefore, we aim to guide the network selectively focus on head regions when estimating the density maps for crowd counting,
 no matter how complex the background is and how various the distributions are.

The attention model has been widely used for different tasks with different focuses,
 e.g. focusing on different patches that relevant to task domain and specific objects for image classification and scene labeling, respectively;
focusing on feature maps with different resolutions for image segmentation;
focusing on different joints and relevant motions for action recognition and focusing on salient features at frame level for video captioning.
For crowd counting, the attention model could be an effective tool to guide the network focusing on head locations,
which are the most important cue for crowd counting.
Therefore, an attention model is introduced to identify how much attention to pay to features at different locations.
Concretely, we use the attention model to concentrate more on head regions, meanwhile suppressing the background regions and body parts in images.

Herein, we briefly introduce the implementation of the attention model used in this work.
Suppose convolutional features in layer i as $f^i$, the soft attention is generated as:

\begin{equation}
\label{eq_AM1}
S = \varphi(W \copyright f^i + b)
\end{equation}
where $\varphi$ is a nonlinear activation function and $\copyright$ denotes convolution operation.
The attention model aims to identify how much attention to pay to features at different locations,
 which could be achieved by generating the probability scores with a softmax operation applied to $S$ spatially:

\begin{equation}
\label{eq_AM2}
M_p = \frac{e^{S_p}} {\sum_{p^{\prime}\in P} e^{S_{p^{\prime}}}}
\end{equation}
$p$ stands for the locations of pixels in soft attention $S$ and $M$ is the probability map.
$M_p$ reflects the probability of presenting head region in position $p$.
By visualizing $M$, we can visualize the attention at different locations,
and the visualization of $M$ is illustrated in Section \ref{sec_attention map}.
Note that $M_p$ is shared across all channels.
The learned probability map is finally multiplied to feature maps in layer $i+1$ to generate attention features, as equation \ref{eq_AM3} shows:

\begin{equation}
\label{eq_AM3}
F^{att} = f^{i+1}\odot M
\end{equation}

Where $\odot$ denotes element-wise product.
Before this operation, the channel of $M$ is expanded as the same as $f^{i+1}$.
$F^{att}$ is the refined attention feature map, which is the feature re-weighted by the probability scores, and has the same size as $f^{i+1}$.

To this end, the trained attention model could adaptively select the relevant positions
 where the heads are located and assigned them higher weights.
This makes the AM-CNN very suitable for crowd counting.

In this work, we generate the probability map from the concatenated multi-scale feature maps.
It may be argued that incorporating attention models into the shallow CNN branches directly is also practical.
Therefore, we tried different architectures with the attention model and will talking about it in section \ref{sec:exper5.1}.

\subsection{Loss Function}
Most of previous methods use Euclidean distance as the loss function for counting task. As equation \ref{eq_distance1} shows:

\begin{equation}
\label{eq_distance1}
L_{ED} =\frac{1}{N}\sum_{i=1}^N (F(X_i,\Theta) - D_i)^2
\end{equation}

Where $N$ is the number of the training samples, $D$ is the ground-truth density map and $F$ is the function that mapping the input $X_i$ to the estimated density map with parameters $\Theta$.
In this paper, Euclidean distance is also selected as the loss function.
Differently, considering that the sizes of the input images are not fixed, the Euclidean distance is divided by the number of pixels $Pix$.

\begin{equation}
\label{eq_loss1}
L_{ED} =\frac{1}{N}\sum_{i=1}^N \frac{(F(X_i,\Theta) - D_i)^2}{Pix_i}
\end{equation}

We also find that for sparse crowd examples, especially the one only contains several persons,
 the Euclidean loss is usually very small, 
 which indicates that these samples receive insufficient treatment during training.
Inspired by \cite{hu2016dense}, we add a relative deviation loss to address this problem.
Hu et al. \cite{hu2016dense} take relative deviation as one of the evaluation criterions but only
use Maximum Excess SubArrays (MESA) distance as counting loss function.
The relative deviation loss used in this paper can be formulated as follows:

\begin{equation}
\label{eq_loss2}
L_{RD} =\frac{1}{N} \sum_{i=1}^N (\frac{y_i - y^{'}_i}{y_i + z})^2
\end{equation}

Where $y_i$ is the ground-truth counts and $y^{'}_i$ is the sum of pixel values of the estimated density map.
$z$ stands for a constant which is used to avoid the errors being divided by zero.
The combination of the 2 loss functions is displayed in equation \ref{eq_loss3}:

\begin{equation}
\label{eq_loss3}
L = L_{ED} + \alpha L_{RD}
\end{equation}

Since the number of pixels in a training sample is usually more than $10^{6}$ and $Pix_i$ is not included in $L_{RD}$,
the loss weight $\alpha$ is set as $10^{-7}$ in the experiments.

\section{Implementation Details}
\label{sec:implementation}
The proposed method is conducted on $3$ highly challenging publicly datasets:
 ShanghaiTech \cite{zhang2016single}, UCF\uline{~}CC\uline{~}50 \cite{idrees2013multi} and WorldExpo'$10$ \cite{zhang2015cross}.
The details of the datasets can be found in Section \ref{sec:experiment}.
We shall firstly describe how to generate ground-truth density maps,
 and then introduce the details of training procedure, which include data process and parameters setting.

\subsection{Density map generation}
The ground-truth density map is converted from the labelled head locations in the original image.
Previous works \cite{zhang2015cross,sindagi2017generating,zhang2016single} generate density maps by locating a Gaussian kernel on the objects.
Zhang et al. \cite{zhang2015cross} sum a 2D Gaussian kernel and a bivariate normal distribution to map the heads and bodies,
but it is only applicable for sparse crowd examples.
Sindagi et al. \cite{sindagi2017generating} use same size of Gaussian kernels for all objects, which cannot illustrate the perspective of the scene.
Similar to \cite{zhang2016single}, we use the geometry-adaptive Gaussian kernels to generate density maps.
Suppose there are $J$ objects in the original image and one of the heads is located in pixel $x_i$,
 then the generation of density map $D$  can be formulated as:

\begin{equation}
\label{eq_density}
D(x) = \sum_{x_i \in J} \mathcal{N}(x - x_i,\sigma_i)
\end{equation}

Where $\mathcal{N}$ is a Gaussian kernel and $\sigma$ represents the variance.
For ShanghaiTech Part\uline{~}A and UCF\uline{~}CC\uline{~}50 datasets,
$\sigma_i$ is computed by k-nearest nighbour (KNN) according to the average distance between the object and its $2$ neighbours.
WorldExpo'$10$ dataset provides perspective maps $\textit{P}$ and $\sigma_i$ is defined as $0.2*$ \textit{$P_i$}.
Since crowds in ShanghaiTech Part\uline{~}B is sparse and perspective maps are not provided, we set $\sigma$ as 4.

\subsection{Training Procedure}
\textbf{Pre-train:} Since some datasets provide limited training images,
 we adopt image cropping for ShanghaiTech and UCF\uline{~}CC\uline{~}50 datasets to expand the training sets.
$C_p$ patches with $1/4$ size of the original image are cropped in random locations to pre-train the shallow CNN branches separately.
Note that the attention model is not included when pre-training the shallow branches.
A convolution operation with $1\times1$ filter is used to generate density map following the former 4 convolutional layers.
$C_p$ is defined as $9$ and $50$ for ShanghaiTech and UCF\uline{~}CC\uline{~}50 datasets, respectively.

\textbf{Fine-tune:} In the fine-tuning procedure, the training dataset is further expanded.
We crop $C_f$ images and flip them, thus totally getting $2 \times C_f$ patches to fine-tune the AM-CNN.
$C_f$ is defined as $100$ and $150$ for ShanghaiTech and UCF\uline{~}CC\uline{~}50, respectively.
WorldExpo'$10$ dataset provides plenty of training images, so we only expand them by flipping the original images to train the AM-CNN.
The $3$ CNN branches are initialized with the pre-trained parameters and the attention model is random initialized with deviation of $0.01$.

\textbf{Parameters setting:} In the training procedure, the learning rate and momentum are set as $10^{-5}$ and 0.9 respectively for Adam optimization.
The batchsize is set as $1$ for training.
All of the experiments are conducted on GeForce GTX TITAN-X.

\section{Experimental Results}
\label{sec:experiment}

This section presents the experimental results on the $3$ public challenging datasets.
For fair comparison, we use $2$ standard metrics for evaluation as other CNN-based counting methods did.
The $2$ metrics are defined as:

\begin{equation}
\label{eq_MSE}
\begin{split}
MAE = \frac{1}{N}\sum_{i\in N} |y_i - y_i^{'}|,\\
MSE = \sqrt{\frac{1}{N} \sum_{i\in N}(y_i - y_i^{'})^2 }
\end{split}
\end{equation}
Where MAE represents mean absolute error and MSE stands for mean squared error, respectively.
$y_i$ is the ground-truth count and $y_i^{'}$ is the estimated count of the AM-CNN for the $i$-th sample.

\subsection{Structural Adjustment based on ShanghaiTech Part\uline{~}A}
\label{sec:exper5.1}

\begin{figure}[!ht]
\centering
\includegraphics[width=3.0in]{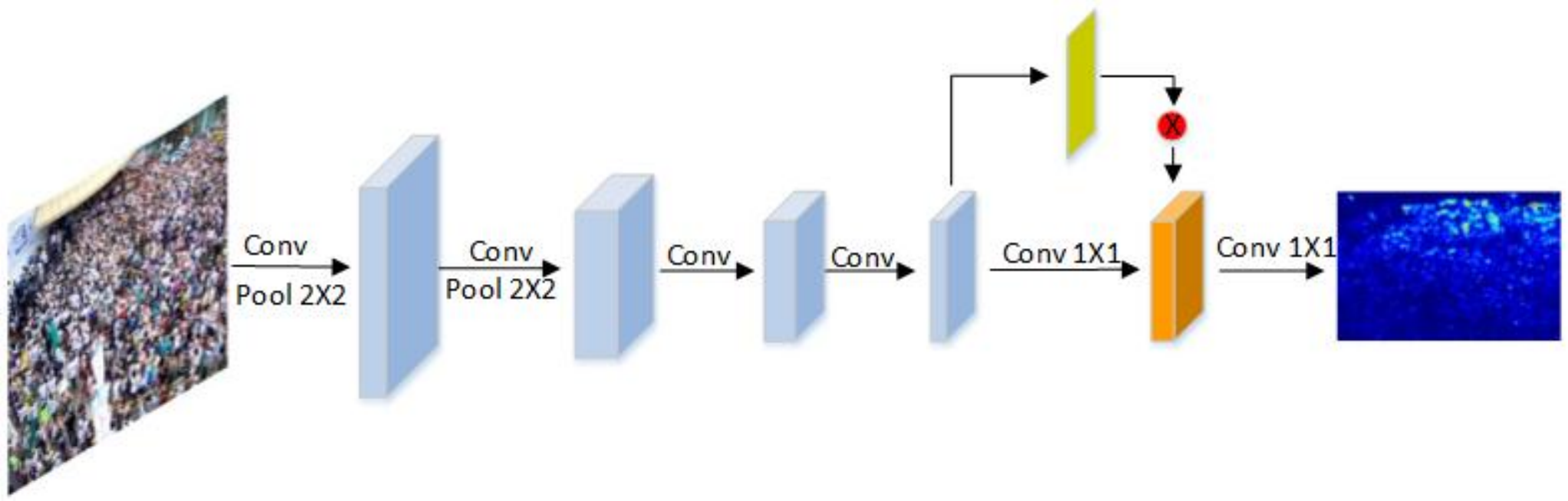}
\caption{Structure of the attention model with a shallow CNN branch.
After 4 convolutional layers, the attention model is utilized to generate attention features directly.
The filter sizes in the former 4 convolutional layers are represented as L (Large, 9-7-7-7), M (Medium, 7-5-5-5) and S (Small, 5-3-3-3).}
\label{fig_1branch}
\end{figure}

\begin{figure}[!ht]
\centering
\includegraphics[width=3.0in]{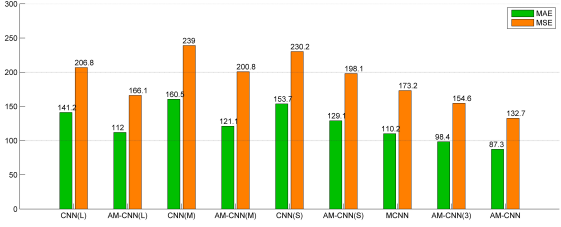}
\caption{Counting results with and without the attention model.
CNN(L), CNN(M) and CNN(S) represent shallow CNN branches displayed in \cite{zhang2016single},
 where L, M and S mean large, medium and small respectively.
 AM-CNN(3) stands for the architecture that integrate the attention model to each shallow CNN branch before feature concatenation.}
\label{fig_result_AMCNN}
\end{figure}

This section presents the effectiveness of the attention model and the structural adjustment of the whole architecture based on ShanghaiTech dataset Part\uline{~}A.
To identify the effectiveness of the attention model, we first incorporated it into a shallow CNN branch.
The incorporation of the attention model and a simple CNN branch is illustrated in Fig. \ref{fig_1branch} and
can be represented as AM-CNN(L), AM-CNN(M) and AM-CNN(S), where L, M and S stand for large, medium and small sizes of convolutional filters.
As the results in Fig. \ref{fig_result_AMCNN} show, 
the counting accuracy increases obviously by using the attention model to emphasize head locations.
The MAEs/MSEs of the AM-CNN(L), AM-CNN(M) and AM-CNN(S) are $112.0/166.1$, $121.1/200.8$ and $129.1/198.1$ while the CNN(L), CNN(M) and CNN(S) get MAEs/MSEs of
$141.2/206.8$, $160.5/239.9$ and $153.7/230.2$, respectively.
The performance improvements achieved by the attention model are $29.2/40.7$(L), $39.4/39.1$(M) and $32.6/29.4$(S), respectively.
These results demonstrate the high effectiveness of the attention model for crowd counting.

We also conducted experiments to determine integrating the attention model before or after feature concatenation.
The first choice is incorporating the attention models to the three shallow CNN branches and then concatenate the attention features for density map generation, named as the AM-CNN(3).
This architecture will generate three different probability maps, each regarding one receptive field.
Another choice is integrating the attention model after the concatenation of the CNN branches, which is the AM-CNN.
Compared with the former, this architecture trains one probability map based on the combination of the multi-scale features,
therefore well exploiting the multi-scale receptive fields when training the attention model.
Results in Fig. \ref{fig_result_AMCNN} show that the second choice achieves higher counting accuracy.
The MAE/MSE of the AM-CNN is $11.1/21.9$ lower than that of the AM-CNN(3).
Besides, compared with the MCNN \cite{zhang2016single}, the proposed method gets a significant performance improvement.
The MAE/MSE of the AM-CNN is $87.3/132.7$ ($22.9/40.5$ lower than that of the MCNN), which also demonstrates the effectiveness of the attention model.

\subsection{Comparison with other CNN-based counting methods}
This section presents the comparison with recent CNN-based methods.
We shall first introduce the details of the datasets, and then discuss the counting results.

\subsubsection{ShanghaiTech}
\label{sec:shanghai}
This dataset was published in \cite{idrees2013multi}, it contains $2$ subsets:
Part\uline{~}A mainly consists of dense crowd examples and Part\uline{~}B mainly focuses on sparse crowd examples.
There are $300$ training images and $182$ testing images in Part\uline{~}A whereas Part\uline{~}B contains $400$ images for training and $316$ for testing.
The crowd density varies greatly in this dataset, making the counting task more challenging than other datasets.
We compare our method with other $5$ recent CNN-based methods in Table \ref{table:shanghai}.

Zhang et al. \cite{zhang2015cross} mainly focus on the cross-scene crowd counting by an operation of candidate scene retrieval.
They retrieve images with similar scenes from training data to fine-tune the trained-CNN for target scene.
In \cite{sindagi2017cnn}, high-level prior is learned by utilizing feature maps trained for density level classification
and thus getting better results than former methods.
On the basis of the MCNN \cite{zhang2016single}, which concatenate feature maps with multi-scale receptive fields,
 Sam et al. \cite{sam2017switching} train a switch-CNN to select a specific CNN regressor for the images.
In addition, they enforced a differential training regimen to tackle the large scale and perspective variations.
Their method improves the performance obviously compared with the MCNN.
Apart from increasing the counting accuracy by adding contextual information,
 Sindagi et al. \cite{sindagi2017generating} use Generative Adversarial Network to sharper the density maps.
Based on the concatenation of multi-scale feature maps, the proposed method exploit an attention model
to emphasize head regions when generating the density map.
In addition, the relative deviation loss compensates small Euclidean distance errors.
For Part\uline{~}A which mainly contains dense crowds, the AM-CNN performs better than other methods expect for the CP-CNN \cite{sindagi2017generating}.
It may result from that the proposed method only uses a density estimator while
 Sindagi et al. \cite{sindagi2017generating} add contextual information which is trained by other two complex structures to their counting architecture.
But the addition of contextual information comes with spurt growth of parameters:
the parameters of the CP-CNN \cite{sindagi2017generating} to be iterated for training an image are about $40$ times than that of the AM-CNN.
Images in Part\uline{~}B mainly focus on sparse crowds, and the proposed AM-CNN gets the state-of-the-art performance on this subset.
Density maps illustrated in Fig. \ref{fig_result_Shanghai} and Fig. \ref{fig_result_World} show that the AM-CNN could focus on every specific head regions in sparse crowds,
which may result in good performance for sparse crowds.
Notably, by integrating an attention model, the proposed method performs much better than the MCNN \cite{zhang2016single}.
The MAEs/MSEs of the AM-CNN (w/o $L_{RD}$) for these 2 sub-sets are $20.6/37.0$ and $10.2/11.5$ lower than that of the MCNN, which demonstrate a significant performance improvement.
The counting accuracy is further increased by adding the relative deviation loss:
The MAE/MSE for Part\uline{~}B reduced by $0.6/3.4$, which is more significant than that for Part\uline{~}A ($2.3/3.5$).
Overall,the attention model guides the network ignore most of the complex backgrounds and give more attention to head regions.
The relative deviation loss relatively expands the estimation errors of sparse crowd examples during training process,
which also plays an important role in crowd counting.

\begin{table}[!ht]
\renewcommand{\arraystretch}{1.3}
\caption{Results on ShanghaiTech dataset}
\label{table:shanghai}
\centering
\begin{tabular}{|p{2.5cm}|p{1cm}|p{1cm}|p{1cm}|p{1cm}|}
\hline
Dataset & \multicolumn{2}{|c|}{Part\uline{~}A} & \multicolumn{2}{|c|}{Part\uline{~}B}\\
\hline
Method & MAE & MSE & MAE & MSE \\
\hline
Cross-Scene \cite{zhang2015cross}   & 181.8 & 277.7 & 32.0 & 49.8\\
\hline
MCNN \cite{zhang2016single}         & 110.2 & 173.2 & 26.4 & 41.3\\
\hline
Cascaded-MLT \cite{sindagi2017cnn}  & 101.3 & 152.4 & 20.0 & 31.1\\
\hline
Switching-CNN \cite{sam2017switching} & 90.4 & 135.0 & 21.6 & 33.4\\
\hline
CP-CNN \cite{sindagi2017generating} & \textbf{73.6} & \textbf{106.4} & 20.1 & 30.1\\
\hline
AM-CNN w/o $L_{RD}$                      & 89.6 & 136.2 & 16.2 & 29.8\\
\hline
AM-CNN with $L_{RD}$                      & 87.3 & 132.7 & \textbf{15.6} & \textbf{26.4}\\
\hline
\end{tabular}
\end{table}

\subsubsection{WorldExpo'10}
\label{sec:world10}
This dataset is the largest one focusing on cross-scene crowd counting.
$199,923$ pedestrians are labelled at their centers of heads, and
$3980$ annotated frames from $1132$ video sequences form the training dataset.
There are totally $103$ scenes captured by $108$ surveillance cameras in this dataset.
Among them, 5 different scenes are used for testing, each consists of $120$ frames and thus forming $5$ sub-sets.
The pedestrian number in the testing set changes significantly over time.
In addition, this dataset provides Region of Interest (ROI) map for each scene.
Referring to \cite{zhang2015cross}, we utilize the ROI map for both training and testing dataset.

\begin{table*}[!ht]
\renewcommand{\arraystretch}{1.3}
\caption{Results on WorldExpo'10 dataset}
\label{table:world}
\centering
\begin{tabular}{|p{2.5cm}|p{1cm}|p{1cm}|p{1cm}|p{1cm}|p{1cm}|p{1cm}|}
\hline
Method & Scene1 & Scene2 & Scene3 & Scene4 & Scene5 & Average \\
\hline
Cross-Scene \cite{zhang2015cross} & 9.8 & 14.1 & 14.3 & 22.2 & 3.7 & 12.9\\
\hline
MCNN \cite{zhang2016single} & 3.4 & 20.6 & 12.9 & 13.0 & 8.1 & 11.6\\
\hline
Switching-CNN \cite{sam2017switching}  & 4.4 & 15.7 & 10.0 & 11.0 & 5.9 & 9.4\\
\hline
CP-CNN \cite{sindagi2017generating} & 2.9 & 14.7 & 10.5 & 10.4 & 5.8 & 8.86\\
\hline
AM-CNN w/o $L_{RD}$ & 3.1 & 13.0 & 9.7 & 10.6 & 5.4 & 8.36\\
\hline
AM-CNN with $L_{RD}$  & \textbf{2.5} & \textbf{13.0} & \textbf{9.7} & \textbf{10.0} & \textbf{4.0} & \textbf{7.84}\\
\hline
\end{tabular}
\end{table*}

Five state-of-the-art algorithms \cite{zhang2015cross}\cite{zhang2016single}\cite{sam2017switching}\cite{sindagi2017generating} which have been introduced
in section \ref{sec:shanghai} are used to compare with the proposed method.
As all of the previous works did, we only display the MAE results in Table \ref{table:world}.
As the results show, 
compared with the MCNN \cite{zhang2016single}, the proposed method achieves a significant improvement by integrating an attention model,
especially for scene 1 and scene 2.
The distributions of people in these 2 scenes change more obviously,
demonstrating that the attention model could emphasize head regions in the image regardless of the non-uniform distribution.
When adding the relative deviation loss, the proposed AM-CNN gets the state-of-the-art results for all subsets.
In scene 1 and scene 5, people distribute more dispersed and the crowds are sparser than other scenes.
The counting accuracy increases more obviously for these 2 subsets, demonstrating that the relative deviation loss plays an important role in sparse crowd counting.
Overall, the proposed method exploits an attention model to focus on head locations,
making the network robust to complex backgrounds and non-uniform distributions.
In addition, the Euclidean loss of sparse crowd examples is usually small, but the relative deviation loss compensates this case.
All the results in table \ref{table:world} demonstrate that the AM-CNN performs well in spite of the cross-scene problem.

\subsubsection{UCF\uline{~}CC\uline{~}50}
This dataset contains $50$ images collected from publicly available web images.
The number of people in one image ranges from $94$ to $4543$ with an average of $1280$.
The scenes in this dataset cover a wide range, such as concerts, stadiums, pilgrimages, protests and marathons.
In the experiment, we perform 5-fold cross validation as other works did.
Images of $1-10$, $11-20$, ... , $41-50$ are used as testing data in the 5 evaluation experiments, respectively.
Table \ref{table:ucf} illustrates the comparison results.

\begin{figure}[!ht]
\centering
\includegraphics[width=3.0in]{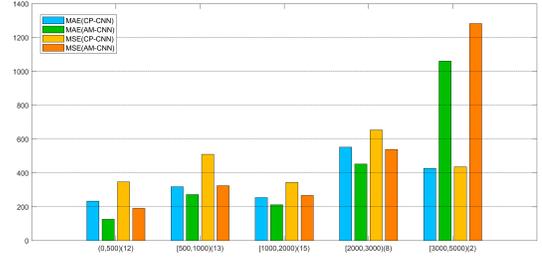}
\caption{Testing Results on UCF\uline{~}CC\uline{~}50 dataset for different densities.
We categorize the density into 5 ranges, which are 0-500, 500-1000, 1000-2000, 2000-3000 and 3000-5000.
The number following the densities in the X-Coordinate is the number of images in that range.}
\label{fig_result_UCF}
\end{figure}

Kumagai et al. \cite{kumagai2017mixture} multiplied appearance-weights output by a gating CNN to a mixture of expert CNNs to address the appearance change problem.
But this method only outputs the number of people while others 
 predict the density map simultaneously.
Authors of \cite{boominathan2016crowdnet} use both deep and shallow CNN branches to extract features from whole image and patches.
They mainly focus on highly dense crowds, but the counting accuracy is not compatible.
Rubio et al. \cite{onoro2016towards} design a Hydra CNN which uses a pyramid of patches as input.
Their scale-aware model does mot need geometric information of scenes.
As Table \ref{table:ucf} shows, the proposed method gets the lowest MAE among these methods.
To explore the performances for different densities,
 we plot a histogram in Fig. \ref{fig_result_UCF} to display the results and the comparisons between the proposed method and the CP-CNN,
 which was the start-of-the art method.
Note that we conduct experiments using the AM-CNN with the relative deviation loss since its effectiveness has been proved on ShanghaiTech and WorldExpo'$10$ datasets.

\begin{table}[!ht]
\renewcommand{\arraystretch}{1.3}
\caption{Results on UCF\uline{~}CC\uline{~}50 dataset}
\label{table:ucf}
\centering
\begin{tabular}{|p{2.5cm}|p{1cm}|p{1cm}|}
\hline
Method & MAE & MSE  \\
\hline
Cross-Scene \cite{zhang2015cross}& 467.0 & 498.5 \\
\hline
Crowdnet \cite{boominathan2016crowdnet}  & 452.5 & --- \\
\hline
MCNN \cite{zhang2016single} & 377.6 & 509.1 \\
\hline
Hydra-CNN \cite{onoro2016towards} & 333.7 & 425.2 \\
\hline
MoCNN \cite{kumagai2017mixture} & 361.7 & 493.3 \\
\hline
Cascaded-MLT \cite{sindagi2017cnn} & 322.8 & 397.9 \\
\hline
Switching-CNN \cite{sam2017switching}  & 318.1 & 439.2\\
\hline
CP-CNN \cite{sindagi2017generating} & 295.8 & \textbf{320.9} \\
\hline
AM-CNN  & \textbf{279.5} & 377.8 \\
\hline
\end{tabular}
\end{table}

\begin{figure*}[!t]
\centering
\includegraphics[width=5.5in]{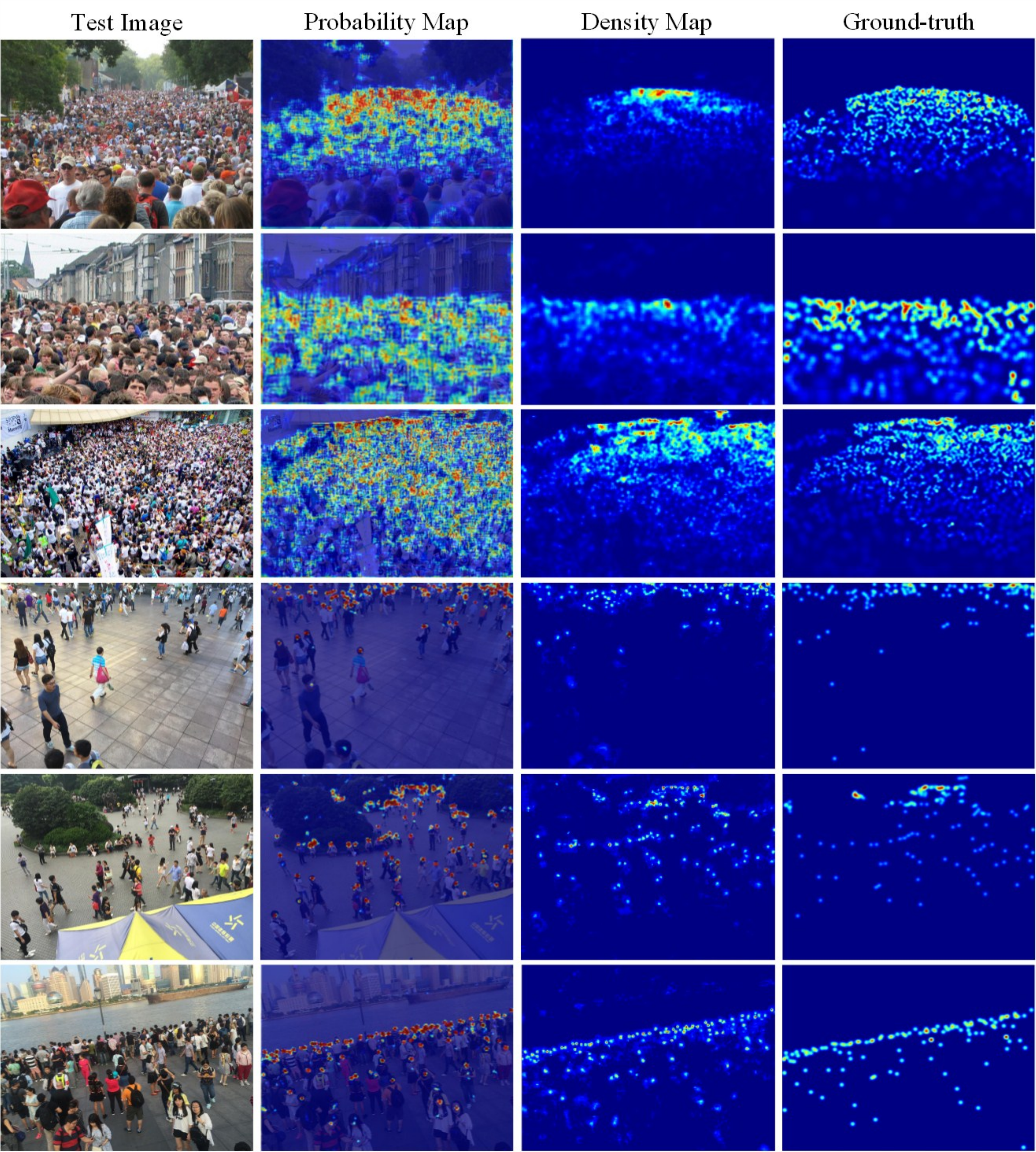}
\caption{Probability and density maps of ShanghaiTech dataset generated by the AM-CNN.
The 3 upper rows are samples selected from Part\uline{~}A and the rest are from Part\uline{~}B.
To illustrate the effectiveness of the attention model concisely, we overlay the probability maps on the original images and set the transparency as 0.7, as the second column shows.
Best viewed in color.}
\label{fig_result_Shanghai}
\end{figure*}

The UCF\uline{~}CC\uline{~}50 is categorized into 5 ranges to show the counting accuracy for different densities.
For scenarios with less than 3000 persons, the AM-CNN performs much better than the CP-CNN. 
However, for extremely dense crowds (with more than $3000$ persons), the AM-CNN performs worse,
 it gets much higher MAE and MSE value compared with the CP-CNN.
It may result from that 
the attention model emphasizes every specific head location in sparse crowds but can only roughly stress the crowd regions of dense crowds.
As aforementioned, the CP-CNN uses two complex structures to exploit contextual information,
and the good performance for extremely dense crowds is at the expense of considerable parameters.
Nevertheless, the AM-CNN can still be applied to many scenarios, such as concerts, stadiums, marathons and markets,
where there are less than $3000$ persons in a single image.

\begin{figure*}[!t]
\centering
\includegraphics[width=5.5in]{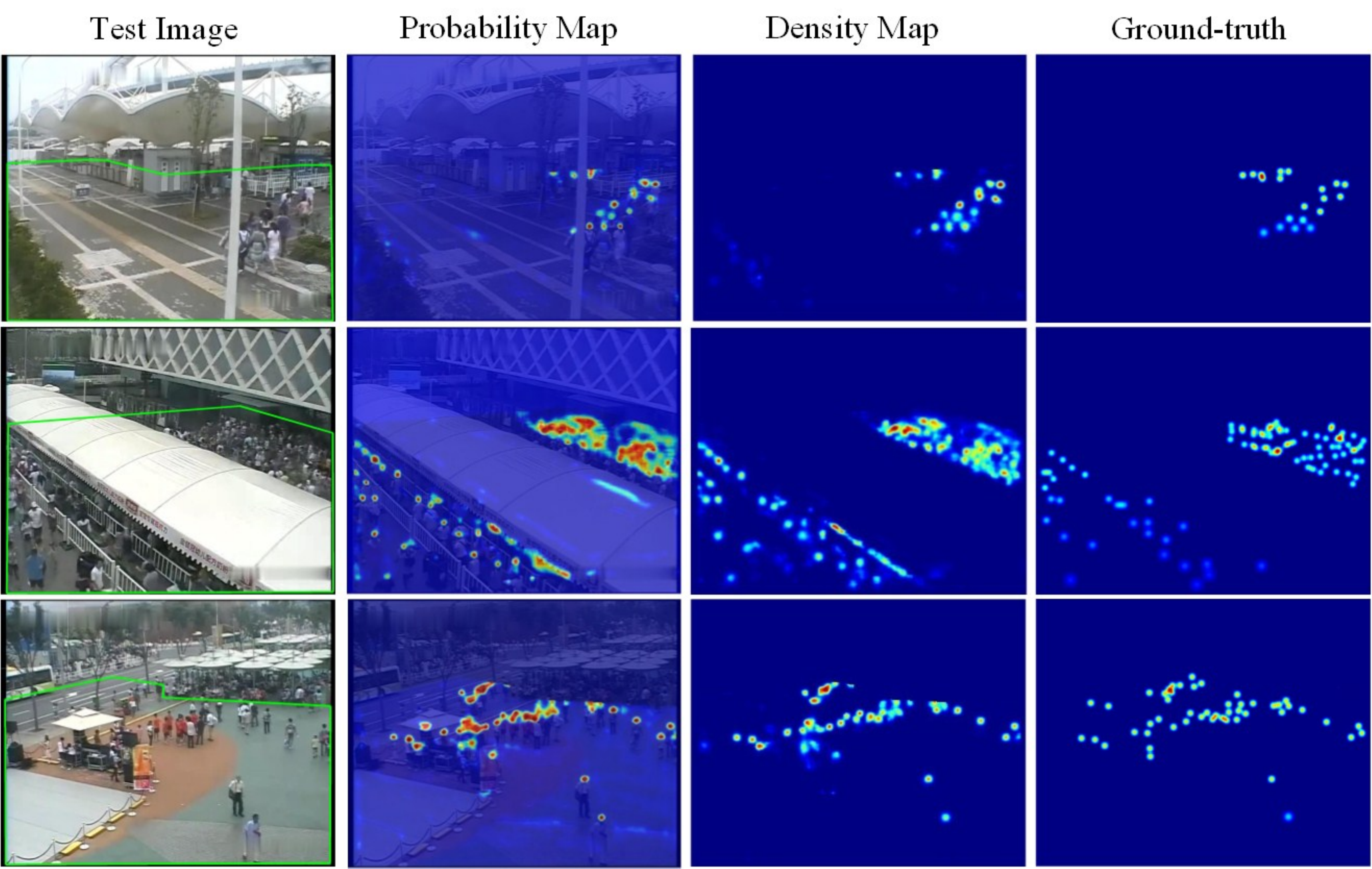}
\caption{Probability and density maps of WorldExpo's 10 generated by the AM-CNN.
Regions in the green shapes are ROIs.
We overlay masks generated according to the ROI on the probability maps and density maps.
Best viewed in color.}
\label{fig_result_World}
\end{figure*}

\subsection{Probability maps}
\label{sec_attention map}

This section displays the probability maps and density maps to explore the influence of the attention model.
Fig. \ref{fig_result_Shanghai} and Fig. \ref{fig_result_World} illustrate representative samples from ShanghaiTech and WorldExpo'10 datasets.
To explore whether the probability maps present higher probability scores in head locations,
we overlay them on the original images.
As Fig. \ref{fig_result_Shanghai} shows, the AM-CNN could concentrate on specific head regions accurately for sparse crowds.
However, for the dense crowds, it can only emphasize the general regions that crowds are located.
It is well known that given an image which contains too many objects to concentrate on,
humans usually focus on the regions where most of the objects are located.
Similarly, it is hard for the attention model to focus on every specific head in a dense crowd, and it concentrates on the region where the crowd is located.


The regions within the green shapes in the first column of Fig. \ref{fig_result_World} are ROIs.
We overlay masks generated according to the ROI on both probability maps and density maps to ignore the masked regions.
Fig. \ref{fig_result_World} demonstrates that the attention model gives much more attention to head locations and thus making the proposed AM-CNN generate clear and accurate density maps.

The probability and density maps displayed in this section demonstrate that the attention model could roughly filtered complex background regions and
body parts before the generation of density maps.
As a result, the density maps become clear and head-focused under the effect of the attention model.

\section{Conclusion}
\label{conclusion}
In this paper, we proposed an attention model convolutional neural network (AM-CNN) to well exploit head locations for crowd counting.
The architecture explicitly gives more attention to head locations and suppresses non-head regions
 by exploiting an attention model to generate a probability map which presents higher probability scores in head regions.
Additionally, a relative deviation loss which plays an important role for sparse crowd density prediction is introduced to compensate the Euclidean loss.
Experiments on three challenging datasets demonstrate the robustness of the AM-CNN to complex backgrounds, scale variations and non-uniform distributions.

\ifCLASSOPTIONcaptionsoff
  \newpage
\fi

\bibliography{mybibfile}
\bibliographystyle{ieeetr}
\end{document}